\pdfoutput=1

\documentclass[11pt]{article}

\usepackage{naacl2021}

\usepackage{times}
\usepackage{latexsym}
\usepackage{booktabs}
\usepackage{graphicx}
\usepackage{amsmath}
\usepackage{float}
\usepackage{multirow}

\usepackage[T1]{fontenc}

\usepackage[utf8]{inputenc}

\usepackage{microtype}

\newenvironment{packed_enum}{
\begin{enumerate}
  \setlength{\itemsep}{1pt}
  \setlength{\parskip}{0pt}
  \setlength{\parsep}{0pt}
}{\end{enumerate}}

\title{Is human scoring the best criteria for summary evaluation?}

\author{Oleg Vasilyev, John Bohannon \\
  Primer Technologies Inc. \\
  San Francisco, California \\
  \texttt{\{oleg,john\}@primer.ai}\\}

\begin{document}
\maketitle
\begin{abstract}
 
Normally, summary quality measures are compared with quality scores produced by human annotators. A higher correlation with human scores is considered to be a fair indicator of a better measure. We discuss observations that cast doubt on this view. We attempt to show a possibility of an alternative indicator. Given a family of measures, we explore a criterion of selecting the best measure not relying on correlations with human scores. Our observations for the BLANC family of measures suggest that the criterion is universal across very different styles of summaries.
\end{abstract}

\section{Introduction}

The goal of summarization is to convey important and only important information of the text in a fluent and comprehensible concise summary, preserving the factual consistency with the text. Almost all of these desired qualities of a summary are subjective to the background and opinion of a reader, arguably except the factual consistency. 

There are several families of automated measures of summary quality. For example, \citet{Saadia2020GoFigure} have classified all evaluation measures into four types: question-answering, text reconstruction, semantic similarity and lexical overlap. Each of these types has families of measures, for example SUM-QE \cite{Stratos2019SUMQE}, APES \cite{Matan2019APES}, Summa-QA \cite{Thomas2019SummaQA} and FEQA \cite{Esin2020FEQA} in question-answering, BLANC-help and BLANC-tune \cite{Oleg2020BLANC} in text reconstruction, BERTScore \cite{Tianyi2020BERTScore}, MoverScore \cite{Wei2020MoverScore} and SUPERT \cite{Yang2020SUPERT} in semantic similarity, ROUGE \cite{Lin2004ROUGE} and Jensen-Shannon \cite{Annie2009Automatically} in lexical overlap.

When it comes to choosing a good evaluation measure, the correlations with human-assigned quality scores is accepted as the crucial criteria. \citet{Saadia2020GoFigure} formulated and explored a framework for judging evaluation measures by correlation with annotated factual errors. Arguably, the factual faithfulness can be annotated objectively, and with detailed classification of factual errors \cite{Wojciech2020Evaluating, Dandan2020What, Oleg2020Sensitivity}. However, other summary qualities are subjective; this forces researchers to be careful in design and usage of human annotations \cite{Manik2020ReEvaluating, Alexander2020SummEval}.

Our motivation to seek criteria alternative or complementary to the correlation with human scores comes from the following observations:
\begin{packed_enum}
\item  Annotation scores are subjective and depend on the types of texts and summaries and on the qualification of annotators. For example, there is a big difference in expert and crowd-sourced scores in \cite{Alexander2020SummEval}\footnote{https://github.com/Yale-LILY/SummEval}.
\item Annotators tend to have a bias favoring anything that helps them to assign a score quickly: extractiveness of the summary, and focus on top of the document \cite{Daniel2020FineTuning}.
\item The annotation itself, as the task of assigning quality scores to a summary by a human, is different from how the summary quality is being valued by a typical human user. A real human reader does not have a goal of scoring a summary, but rather uses the summary to guess the content of the text. 
\end{packed_enum}

In this paper we explore a criterion for selecting an 'optimal' evaluation measure different from maximizing correlation with human scores; we provide evidence that the criterion should be reliably universal across different kinds of summaries. We also observe how a dubious modification of automated evaluation, imitating a human scorer's behavior, can increase correlation with human scores.

\section{Family of measures and max-help criterion}

One of motivations for this exploration is to take a cue from a typical summary user - a user not trying to assign a score to the summary, but rather trying to guess the content of the full text with the help of the summary. In order to imitate such user, the measures based on text reconstruction or on question-answering are the most natural to consider. Following \cite{Oleg2020BLANC}, we consider an evaluation measure as a triplet:
\begin{packed_enum}
\item {\it{Language task}}: Language task to be performed on the text, e.g. text reconstruction or question-answering. The language task is generic, intuitively corresponds to the process of a user understanding the text. The models responsible for the task are trained on large datasets not related to the problem of summarization.
\item {\it{Setup}}: Setup for getting help from the summary. Somehow, the model should get help from the summary, making it easier to perform the language task on the document.
\item {\it{Metrics}}: A specific metrics used to measure the boost in the language task performance, due to the help from the summary. 
\end{packed_enum}

We propose that an optimal measure should on average extract maximal help from the summary. Our reasoning is that the measure most capable in extracting help from summaries should be best fit for quantifying the help. Such a measure would be the most similar to an experienced summary user. Thus, if we have a family of measures, then accordingly to this 'max-help' criterion we should chose a measure that on average (across many samples) outputs a higher value of the boost. 

In this paper we explore the BLANC families, as they leave less ambiguity in the choice of the underlying language model\footnote{https://github.com/PrimerAI/blanc}. Two families defined in \cite{Oleg2020BLANC} differ by the setup. The BLANC-help family gets information from the summary by having the model read the summary before reading and reconstructing the text. The BLANC-tune family gets information from the summary by lightly tuning the model on the summary before reading and reconstructing the text. Practically, the evaluation in both families is arranged to process the text not all at once but sentence-by-sentence.

Measures in each of the families, BLANC-help and BLANC-tune, may differ by the parameters defining the setup, or by the metrics measuring the boost. Several choices of metrics were explored in \cite{Oleg2020Sensitivity}, all giving similar results. 
A choice of the setup parameters also does not make a large difference, except for frequency of masking the text tokens. In this paper we will explore the variations in the setup in both the BLANC-help and BLANC-tune families.

\section{Experiments}
\subsection{Universal trends}

The max-help criterion, formulated in previous section, may be credible only if it does not depend too strongly on the types of texts and summaries. 

In order to excessively verify this assumption, we considered four types of summaries (and the corresponding texts): 
\begin{packed_enum}
  \item CNN summaries from the CNN / Daily Mail dataset \cite{Karl2015CNNDM}.
  \item Daily Mail summaries from the CNN / Daily Mail dataset.
  \item Top two sentences from random daily news.
  \item Random two sentences from random daily news.
\end{packed_enum}
The random daily news were selected as three random documents per day over one year, with the 'summaries'  of the document being two top and two random sentences. We used 1000 samples for each of the four types of summaries.

For BLANC-help family, we found that for all four datasets the optimal or near-optimal setup (accordingly to the max-help criterion) happens to be at: 
\begin{packed_enum}
  \item Interval between masking locations in the text: $gap = 2$.
  \item Number of tokens allowed to be masked at each masking location: $gap\_mask = 1$.
  \item Minimal length of one-word token allowed to be masked is 6 characters: $L_{normal}=6$.
  \item Minimal length of leading token of a composite word is 1 character, i.e. always masked: $L_{lead}=1$.
  \item Minimal length of any of follow-up tokens of a composite word is 1 character, i.e. always masked: $L_{follow}=1$.
\end{packed_enum}

It makes sense that a normal word expressed in BERT model dictionary by single token is supposedly too common to be masked (unless it is a long enough word). This setup is almost the same as the parameters found in \cite{Oleg2020Sensitivity} to maximise correlation with human scores, except $L_{normal}=4$ and $L_{follow}=100$ (follow-up tokens are never masked). Ignoring small effects of the $L$ tokens thresholds, maximizing correlations in this case also maximises average BLANC-help value, as was noticed in \cite{Oleg2020Sensitivity}. As we show here, such lucky coincidence is not a rule: the "max-help" and the "max-human" (maximal correlation with human scores) measures do not always coincide.

\begin{figure}[th]
\includegraphics[width=0.49\textwidth]{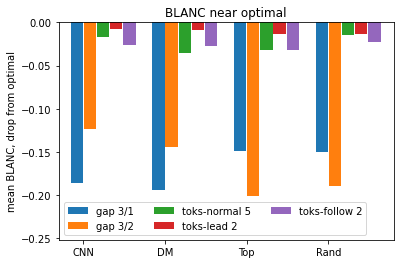}
\caption{Drop of mean BLANC-help value when parameters differ from optimal. The drop is shown as a fraction of the optimal mean BLANC value. The summaries probed are: CNN and DM (from the CNN/Daily Mail dataset), Top and Rand (top two sentences and random two sentences from random news articles). The parameters probed are: 'gap 3/1' is $gap=3$ and $gap\_mask=1$; 'gap 3/2' is $gap=3$ and $gap\_mask=2$; 'toks-normal 5' is $L_{normal}=5$; 'toks-lead 2' is $L_{lead}=2$; 'toks-follow 2' is $L_{follow}=2$.}
\label{fig:BLANChelp_uni}
\end{figure}

The setup may be arranged differently, and may be defined to depend on different parameters. But the question we ask is fundamental for any family of measures: does the 'optimal' max-help evaluation measure remains optimal (or at least near-optimal) for different kinds of texts and summaries? Figure \ref{fig:BLANChelp_uni} provides convincing evidence for the positive answer.

In Figure \ref{fig:BLANChelp_uni} we consider average BLANC-help value obtained with supposedly sub-optimal (different from max-help) setup. We consider a change of $gap$ and $gap\_mask$ to enforce a less frequent and a more frequent masking, and a change in the token length thresholds for masking tokens. Remarkably, the average BLANC-help value drops in each case for all four datasets. The token length thresholds have almost no influence, making a drop just a few percents. Change in frequency of masking has a larger effect, leading to a drop 10\%-20\%.

For BLANC-tune family, we found that for all four datasets the max-help setup happens to be at: 
\begin{packed_enum}
  \item Interval between masking locations in the text for inference: $gap = 3$.
  \item Number of tokens allowed to be masked at each masking location for inference: $gap\_mask = 2$.
  \item The masking at tuning is not random but done 'evenly', the same way as for inference. 
  \item Interval between masking locations in the text for tuning: $gap_{tune} = 4$.
  \item Number of tokens allowed to be masked at each masking location for tuning: $gap\_mask_{tune} = 3$.
  \item Minimal length of one-word token allowed to be masked is 6 characters: $L_{normal}=6$.
  \item Minimal length of leading token of a composite word is 1 character, i.e. always masked: $L_{lead}=1$.
  \item Minimal length of any of follow-up tokens of a composite word is 1 character, i.e. always masked: $L_{follow}=1$.
  \item Probability of replacement of a masked token by another random token at tuning is zero: $p_{replace}=0$. 
  \item Probability of leaving a masked token as it is at tuning is 0.1: $p_{keep}=0.1$.
\end{packed_enum}

Notice that $p_{replace}=0$ differs from the standard BERT training which is done with both $p_{replace}$ and $p_{keep}$ equal 0.1. However, both these probabilities have only weak influence on the BLANC-tune.

\begin{figure}[th]
\includegraphics[width=0.49\textwidth]{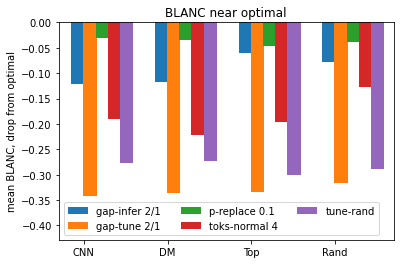}
\caption{Drop of mean BLANC-tune value when parameters differ from optimal. The drop is shown as a fraction of the optimal mean BLANC value. The summaries probed are: CNN and DM (from the CNN/Daily Mail dataset), Top and Rand (top two sentences and random two sentences from random news articles). The parameters probed are: 'gap-infer 2/1' is $gap=2$ and $gap\_mask=1$; 'gap-tune 2/1' is $gap_{tune}=2$ and $gap\_mask_{tune}=1$; 'p-replace 0.1' is $p_{replace}=0.1$; 'toks-normal 4' is $L_{normal}=4$; 'tune-rand' is making tokens masking random rather than even at tuning.}
\label{fig:BLANCtune_uni}
\end{figure}

Figure \ref{fig:BLANCtune_uni} shows several examples of changes of the setup, and again illustrates that the 'optimal' measure remains optimal across all four datasets.

\subsection{Experts and turkers}

If we chose a measure by any criterion that is not optimized by correlation with human scores, then, naturally, such measure would correlate with human score less strongly than the 'max-human' (maximum-correlation) measure of the same family. It is interesting to review how these two measures diverge.

Our "max-help" criterion favors the measures from BLANC-help and BLANC-tune described in the previous section. The "max-human" criterion of maximum correlation with human scores favors somewhat different measures of the same families.

There is only a little difference in BLANC-help "max-human" measure: $L_{normal}=4$; $L_{follow}=100$. The difference in BLANC-tune "max-human" measure is substantial, involving the frequency of masking: $gap = 2$; $gap\_mask = 1$; $gap_{tune} = 2$; $gap\_mask_{tune} = 1$; $L_{normal}=4$; $L_{follow}=100$; $p_{replace}=0.1$. We will consider how the BLANC-tune "max-help" and "max-human" measures diverge.

The "max-help" measure was found using CNN/Daily Mail and random daily news data, and with no need for human scores. There is no need, for that matter, even for human summaries: as shown in the previous section, using sentences from the text leads to the same choice. The "max-human" measure is from \cite{Oleg2020Sensitivity}\footnote{https://github.com/PrimerAI/blanc}. Let see how the measures correlate with human scores of the dataset SummEval \cite{Alexander2020SummEval}\footnote{https://github.com/Yale-LILY/SummEval}.

\begin{table}[htb]
  \begin{center}
    \begin{tabular}{l|l|r|r}
      \textbf{Quality} & \textbf{Correlation} & \textbf{help} & \textbf{human}\\
      \hline
      \multirow{2}{*}{coherence} & Pearson & 0.095 & 0.138\\
      & Spearman & 0.074 & 0.130\\ 
      \hline
      \multirow{2}{*}{consistency} & Pearson & 0.182 & 0.205\\
      & Spearman & 0.170 & 0.198\\ 
      \hline
      \multirow{2}{*}{fluency} & Pearson & 0.149 & 0.152\\
      & Spearman & 0.123 & 0.130\\ 
      \hline
      \multirow{2}{*}{relevance} & Pearson & 0.195 & 0.256\\
      & Spearman & 0.171 & 0.242\\ 
      \hline
    \end{tabular}
    \caption{Correlations of BLANC-tune measures with average expert human scores for the qualities scored in SummEval. Column 'help' shows correlations of maximal-help measure; column 'human' shows correlations of maximal-human measure.}
    \label{tab:max_help_vs_human}
  \end{center}
\end{table} 

Table \ref{tab:max_help_vs_human} shows correlations of both measures with average expert scores assigned to four qualities in \cite{Alexander2020SummEval}. Naturally, the correlations of the max-human measure is higher. But if there is a systematic bias in human scores, and if the max-help criterion has any merit, then we may expect that switching from max-human to max-help would even stronger decrease correlations with non-expert scores, which supposedly might be even further from the max-help 'truth' than the experts. 

Each summary in \cite{Alexander2020SummEval} was scored not only by three experts, but also by five 'turkers' (crowdsource workers). With switch from max-human to max-help measure, the ratio of Pearson correlation with experts to correlation with turkers formally indeed increases by 10\% for relevancy, 70\% for fluency, 68\% for consistency (yet decreases by 1\% for coherence); the correlation with turkers suffers also increase of p-value above 0.05 for all qualities. Similarly, the ratio of Spearman correlation with experts to correlation with turkers increases 15\% for relevance, 47\% for fluency, 77\% consistency (yet decreases 6\% for coherence), and again p-values for turkers increase above 0.05. 

This exercise gives a hope for max-help criterion, or some similar universal principle, not dependent on maximising correlations with human scores.

\subsection{Limited comparison with text}

After reading a summary, an annotator may chose not to review carefully the whole text, but to consider in detail only part of it, whatever attracts attention through a quick glance or a quick read. We can imitate this by using only the most relevant part of the document in calculating BLANC. By most 'relevant' part we mean the part most related to the summary. In modifying BLANC this way, we would supposedly move opposite to the direction described in the previous sections: it is reasonable to expect that correlation with human scores will increase, but this would make a dubious 'improvement' of the BLANC as a measure.

\begin{figure}[th]
\includegraphics[width=0.49\textwidth]{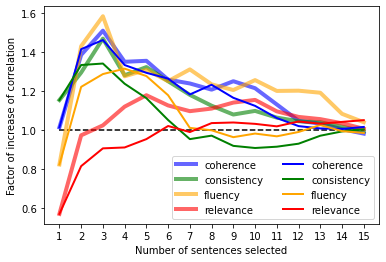}
\caption{Factor by which Spearman correlation of BLANC with human scores increases when only part of text is used for BLANC. The text part is selected as sentences with top BLANC values (thin lines) or as contiguous sentences with highest BLANC (thick lines).}
\label{fig:Restrict_NSentsComb}
\end{figure}

Indeed, it is easy to increase the correlation of BLANC with average expert score for the dataset of 1600 samples of SummEval \cite{Alexander2020SummEval}. We can calculate BLANC separately for each sentence of the text, and select $n$ sentences with highest BLANC. We can consider these selected sentences as the 'text' to deal with, and calculate BLANC on it. Compared to working with full text, Spearman correlation with average expert score increases as shown by thin lines in Figure \ref{fig:Restrict_NSentsComb}. In this and other figures through this section all p-values are below 0.05. 

We can imagine a human expert paying more attention to several (say three or five) most 'promising' sentences of the text. In evaluating relevance, this might be not very different from working with full text. But for other qualities (coherence, consistency, fluency) the correlation increases. 

Naturally, for a human it is easier to review a contiguous piece of text rather than separated pieces, even if this might diminish legitimacy of evaluation of all qualities, including relevance. And, no surprise, BLANC for such contiguous part of text correlates with human scores even better - as shown by thick lines in Figure \ref{fig:Restrict_NSentsComb}.

\begin{figure}[th]
\includegraphics[width=0.49\textwidth]{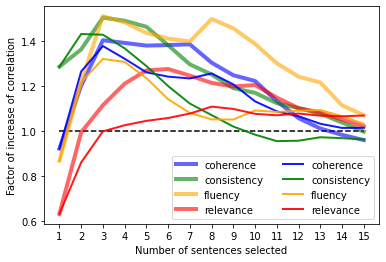}
\caption{Factor by which Spearman correlation of BLANC with human scores increases when only part of text is used for BLANC. The  text part is selected as sentences with top BLANC values (thin lines) or as contiguous sentences having highest average BLANC (thick lines). The resulting BLANC is calculated as average over BLANC of sentences of the selected part of text.}
\label{fig:Restrict_NSentsAvg}
\end{figure}

Figure \ref{fig:Restrict_NSentsAvg} illustrates the same trends when the resulting BLANC is calculated for each selected sentence separately, and then averaged over the sentences.

\begin{figure}[th]
\includegraphics[width=0.49\textwidth]{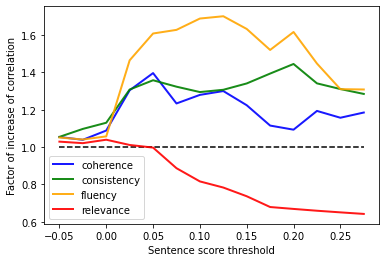}
\caption{Factor by which Spearman correlation of BLANC with human scores increases when only part of text is used for BLANC. The text part is selected as sentences with BLANC exceeding threshold.}
\label{fig:Restrict_ThreshComb}
\end{figure}

Figure \ref{fig:Restrict_ThreshComb} shows the increase of correlations when the text is restricted not by the number of sentences but by a threshold on BLANC of a sentence.

Selection of a part of the text for comparing it with summary is used in SUPERT multi-document evaluation measure \cite{Yang2020SUPERT} as a tool for creating 'reference summary' from each document and then applying evaluation of the summary on the created references. In the context of BLANC here, the selection of a part of the text is done differently and has a clear interpretation: instead of estimating usefulness of the summary in guessing the whole text, we estimate how much the summary would help to guess only the most 'relevant' part of the text. The 'relevant' means the part of the text for which the summary turned out to be most helpful. We suspect that this is equivalent to using only the most promising (for annotator, after reading the summary) part of the text. This does not necessarily mean that the evaluation measure is improved, even though the correlation with human scores is stronger. 

\section{Conclusion}
In this paper, we critically reviewed the assumption that maximal correlation with human scores defines the best evaluation measure for summarization; we provided observations supporting our scepticism. We stated the motivation and made the case for an alternative or at least complementary criterion for choosing an optimal summary evaluation measure from a family of measures. We suggested the maximal average extracted usefulness of summary as such a criterion. We provided observations that the criterion is fairly universal across very different kinds of summaries.

\bibliography{anthology,custom}

\begin{thebibliography}{18}
\expandafter\ifx\csname natexlab\endcsname\relax\def\natexlab#1{#1}\fi

\bibitem[{Bhandari et~al.(2020)Bhandari, Gour, Ashfaq, Liu, and
  Neubig}]{Manik2020ReEvaluating}
Manik Bhandari, Pranav~Narayan Gour, Atabak Ashfaq, Pengfei Liu, and Graham
  Neubig. 2020.
\newblock \href {https://www.aclweb.org/anthology/2020.emnlp-main.751}
  {Re-evaluating evaluation in text summarization}.
\newblock In \emph{Proceedings of the 2020 Conference on Empirical Methods in
  Natural Language Processing}, pages 9347--9359. Association for Computational
  Linguistics.

\bibitem[{Durmus et~al.(2020)Durmus, He, and Diab}]{Esin2020FEQA}
Esin Durmus, He~He, and Mona Diab. 2020.
\newblock \href {https://www.aclweb.org/anthology/2020.acl-main.454} {Feqa: A
  question answering evaluation framework for faithfulness assessment in
  abstractive summarization.}
\newblock In \emph{Proceedings of the 58th Annual Meeting of the Association
  for Computational Linguistics}, pages 5055--5070. Association for
  Computational Linguistics.

\bibitem[{Eyal et~al.(2019)Eyal, Baumel, and Elhadad}]{Matan2019APES}
Matan Eyal, Tal Baumel, and Michael Elhadad. 2019.
\newblock \href {https://www.aclweb.org/anthology/N19-1395} {Question answering
  as an automatic evaluation metric for news article summarization.}
\newblock In \emph{Proceedings of the 2019 Conference of the North American
  Chapter of the Association for Computational Linguistics: Human Language
  Technologies, Volume 1}, pages 3938--3948. Association for Computational
  Linguistics.

\bibitem[{Fabbri et~al.(2020)Fabbri, Kryściński, McCann, Xiong, Socher, and
  Radev}]{Alexander2020SummEval}
Alexander~R. Fabbri, Wojciech Kryściński, Bryan McCann, Caiming Xiong,
  Richard Socher, and Dragomir Radev. 2020.
\newblock \href {https://arxiv.org/abs/2007.12626v3} {Summeval: Re-evaluating
  summarization evaluation.}
\newblock \emph{arXiv}, arXiv:2007.12626v3.

\bibitem[{Gabriel et~al.(2020)Gabriel, Celikyilmaz, Jha, Choi, and
  Gao}]{Saadia2020GoFigure}
Saadia Gabriel, Asli Celikyilmaz, Rahul Jha, Yejin Choi, and Jianfeng Gao.
  2020.
\newblock \href {http://arxiv.org/abs/2010.12834} {Go figure! a meta evaluation
  of factuality in summarization.}
\newblock \emph{arXiv}, arXiv:2010.12834.

\bibitem[{Gao et~al.(2020)Gao, Zhao, and Eger}]{Yang2020SUPERT}
Yang Gao, Wei Zhao, and Steffen Eger. 2020.
\newblock \href {https://www.aclweb.org/anthology/2020.acl-main.124} {Supert:
  Towards new frontiers in unsupervised evaluation metrics for multi-document
  summarization.}
\newblock In \emph{Proceedings of the 58th Annual Meeting of the Association
  for Computational Linguistics}, pages 1347--1354. Association for
  Computational Linguistics.

\bibitem[{Hermann et~al.(2015)Hermann, Kočiský, Grefenstette, Espeholt, Kay,
  Suleyman, and Blunsom}]{Karl2015CNNDM}
Karl~Moritz Hermann, Tomáš Kočiský, Edward Grefenstette, Lasse Espeholt,
  Will Kay, Mustafa Suleyman, and Phil Blunsom. 2015.
\newblock \href
  {https://proceedings.neurips.cc/paper/2015/file/afdec7005cc9f14302cd0474fd0f3c96-Paper.pdf}
  {Teaching machines to read and comprehend.}
\newblock In \emph{Advances in Neural Information Processing Systems},
  volume~28, pages 1693--1701. Curran Associates, Inc.

\bibitem[{Huang et~al.(2020)Huang, Cui, Yang, Bao, Wang, Xie, and
  Zhang}]{Dandan2020What}
Dandan Huang, Leyang Cui, Sen Yang, Guangsheng Bao, Kun Wang, Jun Xie, and Yue
  Zhang. 2020.
\newblock \href {https://www.aclweb.org/anthology/2020.emnlp-main.33} {What
  have we achieved on text summarization?}
\newblock In \emph{Proceedings of the 2020 Conference on Empirical Methods in
  Natural Language Processing}, pages 446--469. Association for Computational
  Linguistics.

\bibitem[{Kryscinski et~al.(2020)Kryscinski, McCann, Xiong, and
  Socher}]{Wojciech2020Evaluating}
Wojciech Kryscinski, Bryan McCann, Caiming Xiong, and Richard Socher. 2020.
\newblock \href {https://www.aclweb.org/anthology/2020.emnlp-main.750}
  {Evaluating the factual consistency of abstractive text summarization.}
\newblock In \emph{Proceedings of the 2020 Conference on Empirical Methods in
  Natural Language Processing}, pages 9332--9346. Association for Computational
  Linguistics.

\bibitem[{Lin(2004)}]{Lin2004ROUGE}
Chin-Yew Lin. 2004.
\newblock \href {https://www.aclweb.org/anthology/W04-1013} {Rouge: A package
  for automatic evaluation of summaries.}
\newblock In \emph{Proceedings of Workshop on Text Summarization Branches Out},
  pages 74--81. Association for Computational Linguistics.

\bibitem[{Louis and Nenkova(2009)}]{Annie2009Automatically}
Annie Louis and Ani Nenkova. 2009.
\newblock \href {https://www.aclweb.org/anthology/D09-1032} {Automatically
  evaluating content selection in summarization without human models.}
\newblock In \emph{Proceedings of the 2009 Conference on Empirical Methods in
  Natural Language Processing}, pages 306--314. Association for Computational
  Linguistics.

\bibitem[{Scialom et~al.(2019)Scialom, Lamprier, Piwowarski, and
  Staiano}]{Thomas2019SummaQA}
Thomas Scialom, Sylvain Lamprier, Benjamin Piwowarski, and Jacopo Staiano.
  2019.
\newblock \href {https://www.aclweb.org/anthology/D19-1320} {Answers unite!
  unsupervised metrics for reinforced summarization models.}
\newblock In \emph{Proceedings of the 2019 Conference on Empirical Methods in
  Natural Language Processing and the 9th International Joint Conference on
  Natural Language Processing}, pages 3246--3256. Association for Computational
  Linguistics.

\bibitem[{Vasilyev et~al.(2020{\natexlab{a}})Vasilyev, Dharnidharka, and
  Bohannon}]{Oleg2020BLANC}
Oleg Vasilyev, Vedant Dharnidharka, and John Bohannon. 2020{\natexlab{a}}.
\newblock \href {https://www.aclweb.org/anthology/2020.eval4nlp-1.2} {Fill in
  the blanc: Human-free quality estimation of document summaries.}
\newblock In \emph{Proceedings of the First Workshop on Evaluation and
  Comparison of NLP Systems}, pages 11--20. Association for Computational
  Linguistics.

\bibitem[{Vasilyev et~al.(2020{\natexlab{b}})Vasilyev, Dharnidharka, Egan,
  Chambliss, and Bohannon}]{Oleg2020Sensitivity}
Oleg Vasilyev, Vedant Dharnidharka, Nicholas Egan, Charlene Chambliss, and John
  Bohannon. 2020{\natexlab{b}}.
\newblock \href {https://arxiv.org/abs/2010.06716} {Sensitivity of blanc to
  human-scored qualities of text summaries.}
\newblock \emph{arXiv}, arXiv:2010.06716.

\bibitem[{Xenouleas et~al.(2019)Xenouleas, Malakasiotis, Apidianaki, and
  Androutsopoulos}]{Stratos2019SUMQE}
Stratos Xenouleas, Prodromos Malakasiotis, Marianna Apidianaki, and Ion
  Androutsopoulos. 2019.
\newblock \href {https://www.aclweb.org/anthology/D19-1618} {Sum-qe: a
  bert-based summary quality estimation model.}
\newblock In \emph{Proceedings of the 2019 Conference on Empirical Methods in
  Natural Language Processing and the 9th International Joint Conference on
  Natural Language Processing}, pages 6005--6011. Association for Computational
  Linguistics.

\bibitem[{Zhang et~al.(2020)Zhang, Kishore, Wu, Weinberger, and
  Artzi}]{Tianyi2020BERTScore}
Tianyi Zhang, Varsha Kishore, Felix Wu, Kilian~Q. Weinberger, and Yoav Artzi.
  2020.
\newblock \href {http://arxiv.org/abs/1904.09675v3} {Bertscore: Evaluating text
  generation with bert.}
\newblock \emph{arXiv}, arXiv:1904.09675v3.

\bibitem[{Zhao et~al.(2019)Zhao, Peyrard, Liu, Gao, Meyer, and
  Eger}]{Wei2020MoverScore}
Wei Zhao, Maxime Peyrard, Fei Liu, Yang Gao, Christian~M. Meyer, and Steffen
  Eger. 2019.
\newblock \href {https://www.aclweb.org/anthology/D19-1053} {Moverscore: Text
  generation evaluating with contextualized embeddings and earth mover
  distance.}
\newblock In \emph{Proceedings of the 2019 Conference on Empirical Methods in
  Natural Language Processing and the 9th International Joint Conference on
  Natural Language Processing}, pages 563--578. Association for Computational
  Linguistics.

\bibitem[{Ziegler et~al.(2020)Ziegler, Stiennon, Wu, Brown, Radford, Amodei,
  Christiano, and Irving}]{Daniel2020FineTuning}
Daniel~M. Ziegler, Nisan Stiennon, Jeffrey Wu, Tom~B. Brown, Alec Radford,
  Dario Amodei, Paul Christiano, and Geoffrey Irving. 2020.
\newblock \href {http://arxiv.org/abs/1909.08593v2} {Fine-tuning language
  models from human preferences.}
\newblock \emph{arXiv}, arXiv:1909.08593v2.

\end{thebibliography}
\bibliographystyle{acl_natbib}

\end{document}